# Periodicity Extraction using Superposition of Distance Matching Function and One-dimensional Haar Wavelet Transform

**Dr. V. Asha**
Department of Computer Applications,
New Horizon College of Engineering
Bangalore,
Karnataka, India
*v_asha@live.com*

**Dr. N.U. Bhajantri**
Department of Computer Science & Engineering,
Government Engineering College,
Chamarajanagar, Mysore District,
Karnataka, India
*bhajan3nu@gmail.com*

**Dr. P. Nagabhushan**
Department of Studies in Computer Science,
University of Mysore,
Mysore, Karnataka, India
*pnagabhushan@hotmail.com*

**Abstract**

Periodicity of a texture is one of the important visual characteristics and is often used as a measure for textural discrimination at the structural level. Knowledge about periodicity of a texture is very essential in the field of texture synthesis and texture compression and also in the design of frieze and wall papers. In this paper, we propose a method of periodicity extraction from noisy images based on superposition of distance matching function (DMF) and wavelet decomposition without de-noising the test images. Overall DMFs are subjected to single-level Haar wavelet decomposition to obtain approximate and detailed coefficients. Extracted coefficients help in determination of periodicities in row and column directions. We illustrate the usefulness and the effectiveness of the proposed method in a texture synthesis application.

## 1. Introduction

Regular textures have periodic patterns where size, shape and intensity of all texture elements are repeating in equal intervals along two independent directions. According to Liu et al., a perfectly regular texture is a regular congruent tiling of a specific texture element [1]. Near-regular textures are considered to be statistical departures from regular textures [1] and are often found in textile garments with different periodic patterns [2]. Analyzing a scanned picture containing patterns is equivalent to analyzing a textile image captured through camera as is the case in textile inspection. This method will be much robust due to the fact that when the method can work well for scanned images that are with more additive noise it will work better for textile images captured through camera that have comparatively less noise. Though methods involving Fourier transform [3], autocorrelation [4], Reney's entropy [5,6] and co-occurrence matrix properties like inertia [7] and distance matching function [8] are available for periodicity determination, we intend to choose superposition of distance matching function (DMF) proposed in [9] along with Haar wavelet for extracting periodicities. Method of superposition of DMF for periodicity determination is superior over other method due to the fact that the method can be employed for extraction of square or rectangular periodic pattern from regular/near-regular/noisy textures of size ranging from finite to infinite. Size of a periodic pattern is defined as $P_C \times P_R$ where $P_C$ is the periodicity along column direction (i.e., number of rows in a periodic pattern) and $P_R$ is the periodicity along row direction (i.e., number of columns in a periodic pattern). Textures are made of repetition of fundamental elements called texels or texture elements [10]. If the size of the periodic pattern is same as the size of texel, then the texture is said to contain simple patterns. If the size of the periodic pattern is more than the size of texel, then the texture is said to contain complex patterns (such as staggered, chessboard and compound patterns) where the periodic pattern has more than a texel [11]. Texture with complex patterns can have texels of different type also. Texel identification is very much useful in the field of image processing, computer vision, and computer graphics such as texture synthesis and texture compression [6,7,8]. Superposition of DMF for periodicity extraction in row / column direction involves summing up of row / column DMFs [9]. This method can tolerate noise to some extent. Since the level of additive noise in poor quality image (such as scanned image) is high, the summed up DMF itself will be noisy due to additive noise occurring during scanning. As the periodicity extraction is based on occurrence of deep valley in the summed-up DMF plot, the presence of noisy affects the periodicity determination. Hence, the noisy summed-up DMF is subjected to single level wavelet decomposition to get approximate and detailed coefficients. Looking at the deep valleys in the approximate coefficients, one can extract the periodicity without de-noising the input textures. The organization of the paper is as follows: Section 2 presents a brief review on Superposition of DMF for periodicity determination proposed in [9] and the impact of noise on it for scanned image containing additive noise. Section 3 describes the proposed model for periodicity determination. Section 4 illustrates the usefulness and the effectiveness of the proposed method for texture synthesis. Section 5 has the conclusions.

**2. Brief review on Superposition of DMF**

The one-dimensional DMF, $d(\Delta)$, of a row or column vector, $g$, of size, $N$, pixels can be defined as [8]

$$d(\Delta) = \sum_{i=0}^{N-\Delta-1} [g(i) - g(i+\Delta)]^2 \qquad (1)$$

where $\Delta$ refers to the number of pixels in the row or column direction. In order to obtain the periodicity in a row or column direction, $d(\Delta)$ can be calculated for different values of $\Delta$ from 1 to $N$-1. Using superposition principle [9], the summed-up value of DMF for row can be calculated using the equation,

$$D_r(\Delta) = \sum_{i=0}^{M-1} \left( \sum_{i=0}^{N-\Delta-1} [f(r,i) - f(r,i+\Delta)]^2 \right) \qquad (2)$$

Similarly, the summed-up value of DMF for column can be calculated using the equation [9],

$$D_c(\Delta) = \sum_{i=0}^{N-1} \left( \sum_{i=0}^{M-\Delta-1} [f(i,c) - f(i+\Delta,c)]^2 \right) \qquad (3)$$

These summed up DMF plots have a self-similar pattern at reducing scale for each period. Looking at the occurrence of deep valey and its location, one can easily extract the periodicity from the summed-up DMF plot [9]. This is illustrated using an example. Let $g$ be a 3-bit image function (with gray values between 0 and 7) of size $8 \times 20$ pixels as shown in Table-1 and Fig. 1.

Table-1: Gray values of the 3-bit image

| 6 | 7 | 0 | 1 | 2 | 6 | 7 | 0 | 1 | 2 | 6 | 7 | 0 | 1 | 2 | 6 | 7 | 0 | 1 | 2 |
|---|---|---|---|---|---|---|---|---|---|---|---|---|---|---|---|---|---|---|---|
| 1 | 2 | 4 | 5 | 0 | 1 | 2 | 4 | 5 | 0 | 1 | 2 | 4 | 5 | 0 | 1 | 2 | 4 | 5 | 0 |
| 5 | 6 | 7 | 0 | 1 | 5 | 6 | 7 | 0 | 1 | 5 | 6 | 7 | 0 | 1 | 5 | 6 | 7 | 0 | 1 |
| 7 | 6 | 5 | 4 | 3 | 7 | 6 | 5 | 4 | 3 | 7 | 6 | 5 | 4 | 3 | 7 | 6 | 5 | 4 | 3 |
| 6 | 7 | 0 | 1 | 2 | 6 | 7 | 0 | 1 | 2 | 6 | 7 | 0 | 1 | 2 | 6 | 7 | 0 | 1 | 2 |
| 1 | 2 | 4 | 5 | 0 | 1 | 2 | 4 | 5 | 0 | 1 | 2 | 4 | 5 | 0 | 1 | 2 | 4 | 5 | 0 |
| 5 | 6 | 7 | 0 | 1 | 5 | 6 | 7 | 0 | 1 | 5 | 6 | 7 | 0 | 1 | 5 | 6 | 7 | 0 | 1 |
| 7 | 6 | 5 | 4 | 3 | 7 | 6 | 5 | 4 | 3 | 7 | 6 | 5 | 4 | 3 | 7 | 6 | 5 | 4 | 3 |

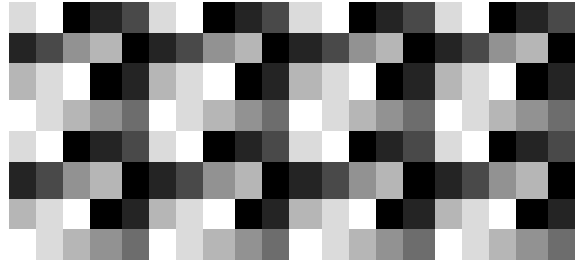

**Fig. 1. Image *g* shown in enlarged form with each pixel represented as a square for clarity**

Let $g_c$ be the function formed from the function $g$ after corrupting it with noisy pixels shown bold as in Table-2 and Fig. 2.

Table-2: Gray values of the 3-bit image with noisy pixels

| 6 | 7 | 0 | 1 | **5** | 6 | 7 | 0 | 1 | 2 | 6 | 7 | 0 | 1 | 2 | 6 | 7 | 0 | 1 | 2 |
|---|---|---|---|---|---|---|---|---|---|---|---|---|---|---|---|---|---|---|---|
| 1 | 2 | 4 | 5 | 0 | 1 | 2 | 4 | 5 | 0 | 1 | 2 | 4 | 5 | 0 | 1 | 2 | 4 | 5 | 0 |
| 5 | 6 | 7 | 0 | 1 | 5 | 6 | 7 | **7** | 1 | 5 | 6 | 7 | 0 | 1 | 5 | 6 | 7 | 0 | 1 |
| 7 | 6 | 5 | 4 | 3 | 7 | 6 | 5 | 4 | 3 | 7 | 6 | **0** | 4 | 3 | 7 | 6 | 5 | 4 | 3 |
| 6 | 7 | 0 | 1 | 2 | 6 | 7 | 0 | 1 | 2 | 6 | 7 | 0 | 1 | 2 | 6 | 7 | 0 | 1 | 2 |
| 1 | 2 | 4 | 5 | 0 | 1 | 2 | 4 | 5 | 0 | 1 | 2 | 4 | 5 | 0 | 1 | 2 | 4 | 5 | 0 |
| 5 | 6 | 7 | 0 | 1 | 5 | 6 | 7 | 0 | 1 | 5 | **1** | 7 | 0 | 1 | 5 | 6 | 7 | 0 | 1 |
| 7 | 6 | 5 | **3** | 3 | 7 | 6 | 5 | 4 | 3 | 7 | 6 | 5 | 4 | 3 | 7 | 6 | 5 | 4 | 3 |

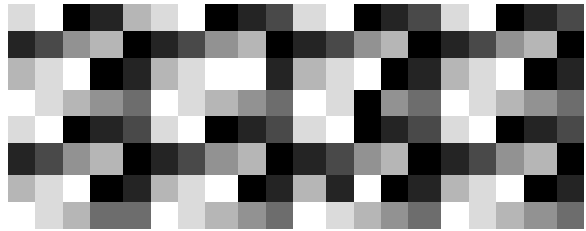

**Fig. 2. Image $g_c$ shown in enlarged form with each pixel represented as a square for clarity.**

Table-3: DMF values for all rows for the 3-bit image with noisy pixels

| | | Pixel location along row dimension | | | | | | | | | | | | | | | | | | |
|---|---|---|---|---|---|---|---|---|---|---|---|---|---|---|---|---|---|---|---|---|
| | | 1 | 2 | 3 | 4 | 5 | 6 | 7 | 8 | 9 | 10 | 11 | 12 | 13 | 14 | 15 | 16 | 17 | 18 | 19 | 20 |
| Row number | 1 | 0 | 256 | 454 | 428 | 220 | 9 | 173 | 307 | 323 | 167 | 9 | 105 | 181 | 197 | 99 | 9 | 52 | 76 | 50 | 16 |
| | 2 | 0 | 127 | 196 | 182 | 97 | 0 | 95 | 142 | 128 | 65 | 0 | 63 | 88 | 74 | 33 | 0 | 31 | 34 | 20 | 1 |
| | 3 | 0 | 242 | 398 | 372 | 206 | 98 | 174 | 272 | 246 | 103 | 49 | 155 | 202 | 176 | 84 | 0 | 52 | 76 | 50 | 16 |
| | 4 | 0 | 114 | 152 | 158 | 126 | 50 | 94 | 122 | 83 | 71 | 25 | 59 | 87 | 48 | 36 | 0 | 4 | 12 | 18 | 16 |
| | 5 | 0 | 256 | 454 | 428 | 220 | 0 | 188 | 328 | 302 | 152 | 0 | 120 | 202 | 176 | 84 | 0 | 52 | 76 | 50 | 16 |
| | 6 | 0 | 127 | 196 | 182 | 97 | 0 | 95 | 142 | 128 | 65 | 0 | 63 | 88 | 74 | 33 | 0 | 31 | 34 | 20 | 1 |
| | 7 | 0 | 306 | 394 | 368 | 270 | 50 | 238 | 268 | 242 | 187 | 25 | 135 | 202 | 176 | 84 | 0 | 52 | 76 | 50 | 16 |
| | 8 | 0 | 66 | 114 | 120 | 79 | 1 | 43 | 79 | 83 | 59 | 1 | 23 | 49 | 53 | 39 | 1 | 3 | 12 | 18 | 16 |

The summed up DMF in row direction is calculated by summing up these values columnwise and is given in Table-4 and in Fig. 3.

Table-4: DMF values for all rows for the 3-bit image with noisy pixels

| | Pixel location along row dimension | | | | | | | | | | | | | | | | | | | |
|---|---|---|---|---|---|---|---|---|---|---|---|---|---|---|---|---|---|---|---|---|
| | 1 | 2 | 3 | 4 | 5 | 6 | 7 | 8 | 9 | 10 | 11 | 12 | 13 | 14 | 15 | 16 | 17 | 18 | 19 | 20 |
| Summed DMF | 0 | 1494 | 2358 | 2238 | 1315 | 208 | 1100 | 1660 | 1535 | 869 | 109 | 723 | 1099 | 974 | 492 | 10 | 277 | 396 | 276 | 98 |

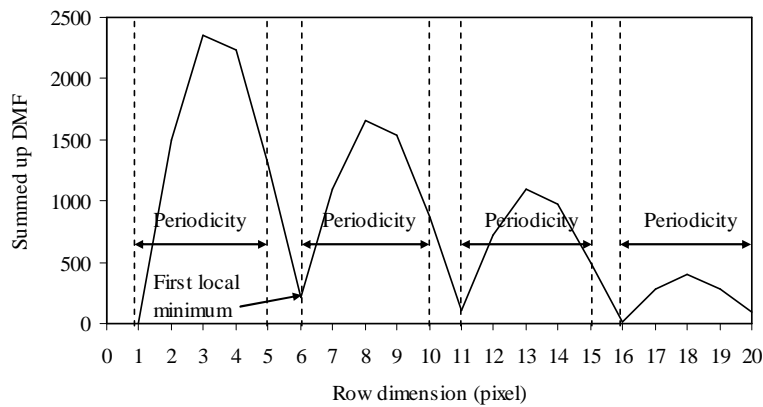

**Fig. 3. Plot of summed up values of row DMF.**

Pixel location along the row direction just before the occurrence of deep valley that yields the first local minimum corresponds to periodicity in that direction. Also, the distance between two successive local minima in row or column direction corresponds to periodicity in that direction. Thus, periodicity for the 3-bit image in row direction is 5 pixels as clearly shown in Fig. 3.

Even for a texture with highly distorted texels of simple patterns as shown in Fig. 4 [9], the superposition of DMF yields very good result as seen from Fig. 5.

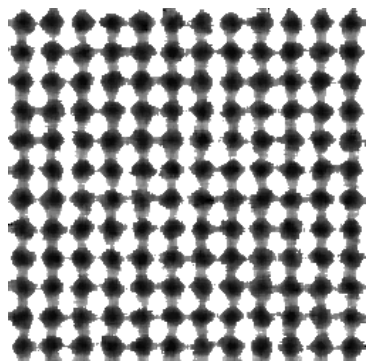

**Fig. 4. A near-regular synthetic texture from [9].**

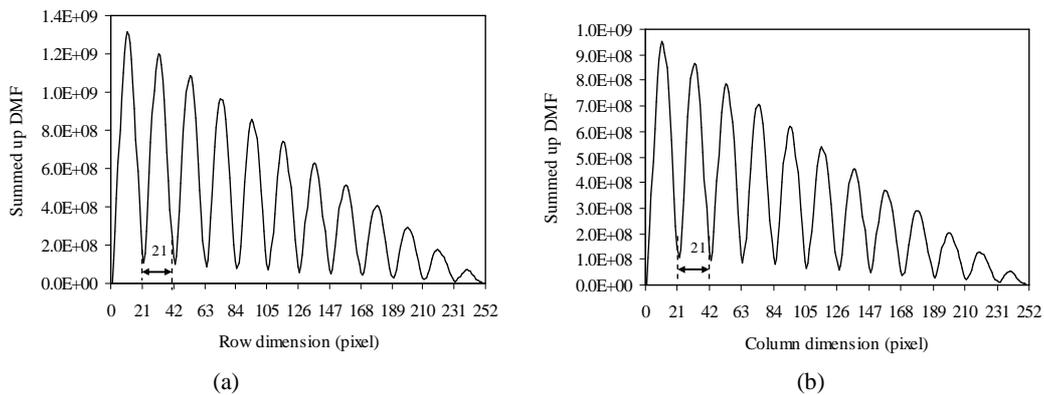

**Fig. 5. Plots of summed-up DMFs for the near-regular synthetic shown in Fig. 5**

Thus, for the texture shown in Fig. 4, both row and column periodicities are found to be 21 pixels and hence the size of periodic pattern (texel size) is $21 \times 21$ pixels as shown in Fig. 6.

In order to study the behaviour of superposition of DMF for a scanned image, image shown in Fig. 6 taken from [12] is used as base image.

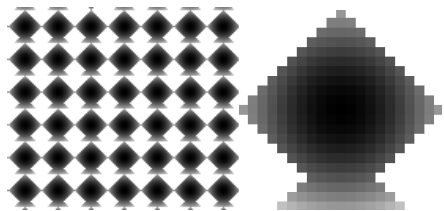

**Fig. 6. Simulated synthetic texture with simple periodic patterns (a) Full view (b) Enlarged view of a periodic pattern wherein each pixel is represented as square for clarity.**

This synthetic texture is printed and scanned using a HP scanner. Fig. 7 shows the scanned image containing additive noise.

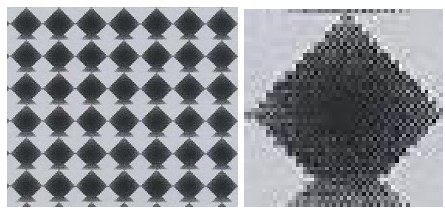

**Fig. 7. Scanned version of the synthetic texture with simple periodic patterns (a) Full view (b) Enlarged view of a periodic pattern**

Comparing the periodic patterns shown in Fig. 6 (b) and Fig. 7 (b), one can easily see the influence of noise on the image quality. Fig. 8 shows the summed up DMFs for the scanned image.

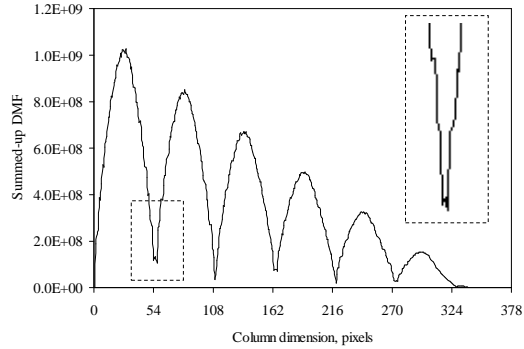

**Fig. 8. Result of superposition of DMF**

The result of superposition of DMF on the scanned image shows that the presence of additive noise in scanned images hinders the occurrence of deep valleys and imparts waviness to the summed-up DMF plot. This DMF plot as a noisy wave needs a denoising technique to smoothen it and extract the deep valleys for periodicity extraction.

**3. Proposed method of periodicity extraction using DMF and Haar wavelet**

Wavelets allow a series of data to be analyzed in multiple resolutions with each resolution representing a different frequency [13]. The wavelet technique takes averages and differences of a signal by splitting the signal into spectrum. Each step in a wavelet transform yields a set of averages and a set of differences (the differences are referred to as wavelet coefficients), wherein each set is half the size of input data. Wavelets are a mathematical tool for hierarchically decomposing functions. They easily allow a function to be described in terms of a coarse overall structure and details that range from broad to very fine. Regardless of whether the function of interest is an image, a curve, or a surface, wavelets offer a sophisticated technique for representing the levels of detail. The DWT of a signal $x$ is calculated by passing it through a series of filters. First the samples are passed through a low pass filter with impulse response $g$ resulting in a convolution of the two as given by

$$y[n] = (x * g)[n] = \sum_{k=-\infty}^{\infty} x[k]g[n-k] \qquad (5)$$

Simultaneously the signal is also decomposed using a high-pass filter *h*. The outputs are the detail coefficients (from the high-pass filter) and approximation coefficients (from the low-pass) given by

$$y_{low}[n] = \sum_{k=-\infty}^{\infty} x[k]g[2n-k] \qquad (6)$$

$$y_{high}[n] = \sum_{k=-\infty}^{\infty} x[k]g[2n+1-k] \qquad (7)$$

Since half the frequencies of the signal are removed, half the samples can be discarded according to Nyquist's rule. The filter outputs are then sub-sampled by 2. This decomposition has halved the time resolution since only half of each filter output characterizes the signal. However, each output has half the frequency band of the input so the frequency resolution has been doubled. This decomposition scheme is shown in Fig. 9.

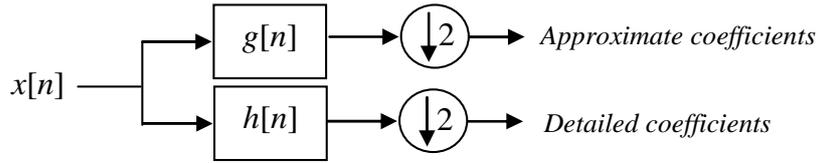

Fig. 9. Decomposition scheme using Haar wavelet

Here, the sub-sampling is denoted by the operator ↓ and the summations can be rewritten as

$$y_{low}[n] = (x * g) \downarrow 2 \qquad (8)$$

$$y_{high}[n] = (x * h) \downarrow 2 \qquad (9)$$

The idea behind the wavelet decomposition on summed-up DMF goes this way. Since the summed-up DMF itself looks like a noisy periodic degenerative signal, the signal is decomposed into two down-sampled signals (approximate coefficients and detailed coefficients). The approximate coefficients represent the low frequency (smoothed) components of the original signal and the detailed coefficients represent the high frequency components of the original signal such as noise. Since the decomposed signals are down-sampled versions of the original sample in terms of approximate and detailed coefficients, the location corresponding to the occurrence of second deep valley is a direct measure of periodicity of the original signal. Thus for the summed up DMF shown in Fig. 8, the result of one-dimensional Haar wavelet decomposition is shown in Fig. 9. The occurrence of second deep valley and the periodic pattern in the approximate coefficients correspond to a periodicity of 54 pixels along row direction. Similar decomposition on summed-up DMF for column direction also yields a periodicity of 54 pixels.

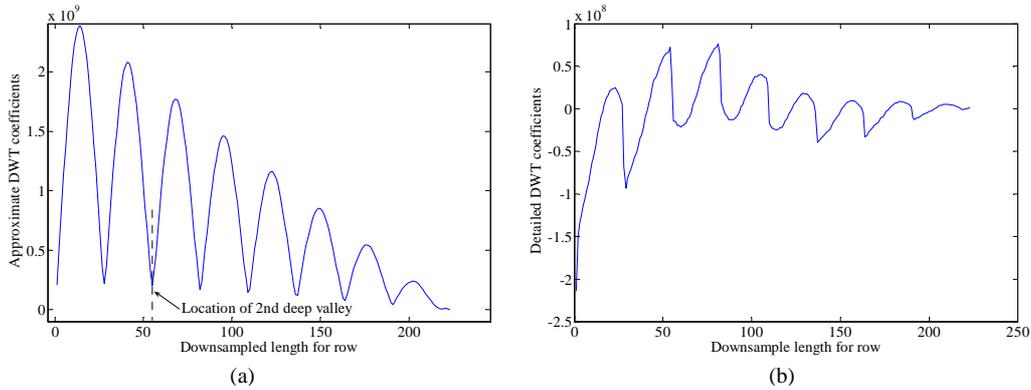

**Fig. 9. Result of 1D discrete wavelet transform on summed up row DMF for the noisy image (a) Approximate coefficients (b) Detailed coefficients.**

## 4. Application to Texture Synthesis

In this section, an application of the periodicity determination is presented for texture synthesis by tiling the appropriate patterns extracted from the texture without breaking the regularity. As far as natural textures are concerned, the major complexity in texture synthesis lies in the extraction of the best periodic pattern that can represent the original texture. An ideal periodic pattern should be the one whose properties are same as those of the texture itself. This can be true for regular synthetic textures wherein all texels are of uniform size, shape and intensity. This cannot be, however, true for natural textures. Hence a comparative study based on both first order and second order statistical properties is made for all periodic patterns against the texture itself. We try to locate the texels whose first order statistical properties (such as mean, variance, skewness, kurtosis, energy and entropy) and co-occurrence matrix properties (such as energy, entropy, contrast, homogeneity and correlation) calculated at unit pixel displacement and different angles are close to those of the texture. Though there are few patterns whose statistical properties are close to those of the entire texture, the best pattern is estimated based on visual appearance of the synthesized texture which preserves the regularity upon tiling. Fig. 10 shows the D20 texture (taken from Brodatz album [14]), wherein the periodic patterns with properties close to those of the entire texture are highlighted. It may be noted that for simplicity, D20 with complete number of periodic patterns is taken for illustration. Table-5 shows the absolute deviation in statistical properties of few periodic patterns whose properties are close to those of texture. It may be noted that since the results for the test images are found to be less sensitive to the direction of displacement vector, the results are presented only for 0° angle. The effectiveness of the periodicity extraction can be well observed by tiling the extracted pattern to artificially generate the texture. Among the identified appropriate periodic patterns, patterns #6, #8 and #11 yield pleasing result of synthesis as clearly seen from Fig. 11. The result of the proposed texture synthesis method is compared with the texture synthesis result based on Reney's entropy method where the synthesized texture was generated by tiling square window randomly selected from the input texture based on histogram comparison using Kolmogorov-Smirnov statistical test at a significance level of 0.01 [5,6]. Results are in good agreement with Reney's method of texture synthesis.

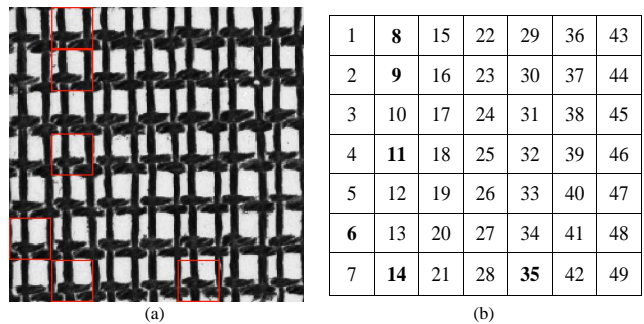

(a)                      (b)

**Fig. 10. Result of statistical comparison for D20 texture of size 231 × 231. (a) D20 with identified appropriate texels. (b) Numbering sequence with identification of appropriate patterns shown in bold.**

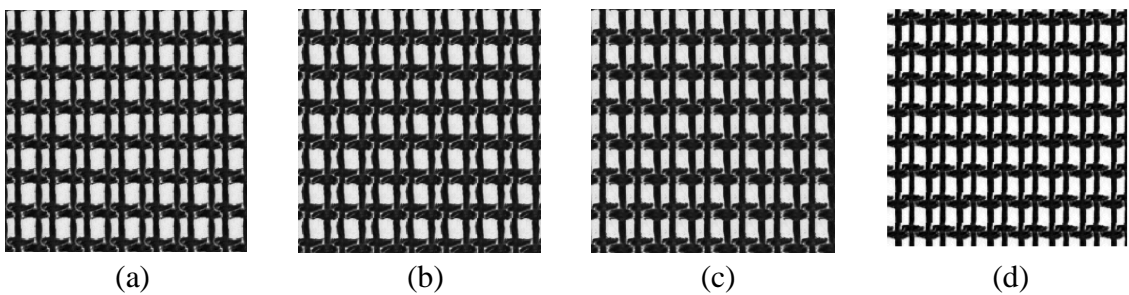

(a)        (b)        (c)        (d)

**Fig. 11. Result of texture synthesis for D20 and comparison with Reney's entropy method. (a) Synthesized texture using texel #6 (b) Synthesized texture using texel #8 (c) Synthesized texture using texel #11 (d) Synthesized texture using Reney's entropy method [13].**

Table-5: Percentage deviation in statistical properties of appropriate texels against the texture D20

| Property | | Texel identity | | | | | |
|---|---|---|---|---|---|---|---|
| | | #6 | #8 | #9 | #11 | #14 | #35 |
| First Order Statistics | Mean | 0.04 | 4.78 | 3.85 | 2.66 | 0.61 | 1.03 |
| | Variance | 0.02 | 2.41 | 3.56 | 2.46 | 3.58 | 1.21 |
| | Skewness | 6.76 | 14.86 | 3.37 | 1.53 | 10.78 | 7.95 |
| | Kurtosis | 1.04 | 0.87 | 4.06 | 2.91 | 8.12 | 3.39 |
| | Energy | 3.27 | 3.17 | 0.82 | 1.67 | 6.92 | 4.80 |
| | Entropy | 0.44 | 2.46 | 1.09 | 1.20 | 0.09 | 0.58 |
| Second Order Statistics | Energy | 14.81 | 15.06 | 15.15 | 15.14 | 14.33 | 13.46 |
| | Entropy | 4.25 | 10.09 | 9.66 | 14.88 | 9.28 | 13.83 |
| | Contrast | 9.18 | 5.56 | 3.99 | 14.31 | 3.37 | 10.44 |
| | Homogeneity | 0.71 | 5.96 | 0.64 | 5.64 | 0.84 | 12.14 |
| | Correlation | 8.09 | 12.94 | 9.48 | 2.36 | 11.32 | 7.95 |

Since the estimation of row periodicity is independent of that of column periodicity, our method of periodicity extraction and texture synthesis (unlike Reney's method) can be applied for textures containing rectangular texels also as the one shown in Fig. 12.

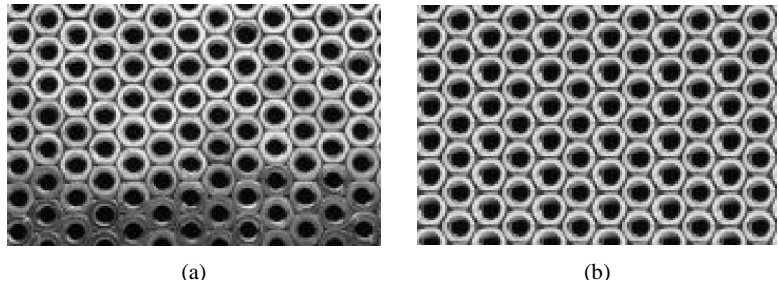

(a)          (b)

**Fig. 12. (a) A near-regular natural texture with rectangular texels (b) Result of texture synthesis**

## 5. Conclusions

In the present work, we have proposed a simple method of extracting periodicity of near-regular textures using superposition of Distance Matching Function and one-dimensional discrete Haar wavelet transform without application of any filtering technique on the input images. The algorithm developed is found to be very efficient for determining periodicities and extracting periodic patterns of square or rectangular shape from near-regular textures. The present method is superior over Reney's method that can extract only square texels. The effectiveness of the proposed method of periodicity extraction is illustrated through texture synthesis application.